# Real-Time Violence Detection Using CNN-LSTM

Mann B. Patel
Charotar University of Science and Technology
18dcs074@charusat.edu.in

*Abstract* - **Violence rates however have been brought down about 57% during the span of past 4 decades yet it doesn't change the way that the demonstration of violence actually happens, unseen by the law. Violence can be mass controlled sometimes by higher authorities, however to hold everything in line one must "Micro-govern" over each movement occurring in every road of each square. To address the butterfly effects impact in our setting, I made a unique model and a theorized system to handle the issue utilizing deep learning. The model takes the input of the CCTV video feeds and after drawing inference, recognizes if a violent movement is going on. And hypothesized architecture aims towards probability driven computation of video feeds and reduces overhead from naively computing for every CCTV video feeds.**

### KEYWORDS

activity recognition, deep learning, inference algorithm, pipeline, supervised, surveillance, violence

### INTRODUCTION

I propose a pseudo real time Violence detection system, which takes a video, may it be with audio or without, and somehow alerts when violent activities are detected. This project solely focuses on taking inferences from whatever data I are able to extract out of the video feeds coming from the CCTV networks to one workstation (or in case of parallelism, cluster). I try to tackle the violent detection challenge using two novel approaches. So, upon doing basic testing, I choose the CNN + LSTM approach, in further cases I also try and test different models of CNN to get which one provides the most accuracy. I also try to extract information from audio of the video, and try get inference from it. Also, I hypothesized the signaling mechanism and convenient algorithm to further a lot computation to a video feed for early detection.

To find out which approach is better, I tried out both approach on, I can just run a simple test run after training over a preprocessed dataset and let them infer over 20 frames randomly extracted from any video on the dataset and see what turns out.

### DATASET

To test our methodology, we work with these three datasets, Hockey Fight Dataset [4], Movies Dataset [5] and Violent-Flows [6]. the 3 datasets captured from closed- circuited-TV, Phone or high-resolution recorder, the quality, number of pixels and length varies between dataset.

• Hockey fights: Dataset composed of equal number of violence and nonviolence action during hockey professional matches, usually Two players participating in close body interaction.

• Movies: This dataset consists fight sequences collected from movies, for the non-violence label - videos of general action activity gathered from movies. The dataset is made up of equal number of violent movie clips and non-violent movie clips. Unlike the Hockey dataset, this dataset varies profoundly between samples.

• Violent-flow: This is a crowd violence dataset. Most of the crowd violence seen in this dataset are clips of football matches.

| Dataset | Description | Total videos | Labelled Violent | Labelled Non-Violent | Final pickled size |
|---|---|---|---|---|---|
| Hockey fights | hockey players | 1000 | 500 | 500 | ≈200 MB |
| Violent-Flows | big crowd videos | 200 | 100 | 100 | ≈100 MB |
| Movies | movies clip | 246 | 123 | 123 | ≈150 MB |

TABLE 1
DATASET SUMMARY

## RELATED WORK

There are a few works on detecting violence more or less via the same both methods as stated above with modifications in the model used. A. Datta , M. Shah et al(2002)[1] proposed a system that infers violence from motion trajectory of the limbs being traced by pose estimation models and then feeding it into an LSTM to get the inference. Tao Zhang, Zhijie Yang, et al (2016)[2] proposed a Gaussian Model of Optical Flows that when passed to the linear classifier gives regions where violence is inferred.

## OBJECTIVE

Our objective is to make a pseudo real time Violence detection system, which takes a video, may it be with audio or without, and somehow alerts when violent activities are detected with highest order of accuracy when model is generalized.

## METHODOLOGY

I try to tackle the violent detection challenge using two base approaches:

- Using Human Pose Estimation + LSTM [1]

- Using Hybrid CNN + LSTM

TABLE I
COMPARISON OF APPROACHES

| Approach | Accuracy | Mean Inference Time |
|---|---|---|
| CNN + LSTM | 92.3% | 742 MS |
| Pose Estimation + LSTM | 87.6 % | 981 MS |

To be noted, for pose estimation I use PoseNET, a TFlite pose estimation model provided by TensorFlow hub, trained on top of mobileNET(Input dimension: 257x257). For the CNN approach I use ResNet50. I compared both methodology and find which one is best for futher adding the audio ensemble to it.

## DATA PREPROCESSING

As a preparation for the input to the graph, few steps were taken in the dataset preparation, initially the videos were sampled to a frame-by-frame sequence as I were limited with computational power. We extract frames from each video into fix batches of frames and then for all perform all possible combination of augmentation methods uniquely on each dataset, like removal of black borders from "Hockey" dataset.

The input to the encoder model is images that are subtraction(pixel by pixel) of adjacent frames .This was done in order to include spatial movement in the input videos instead of the raw pixels from each frame. In Figure 1, we can see only the pushing spatial movement is seen after we subtract the two adjacent frames.

FIGURE 1
FRAME TO FRAME DIFFERENCE

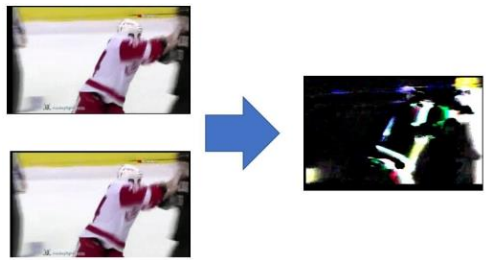

FIGURE 2
DARK EDGES REMOVAL

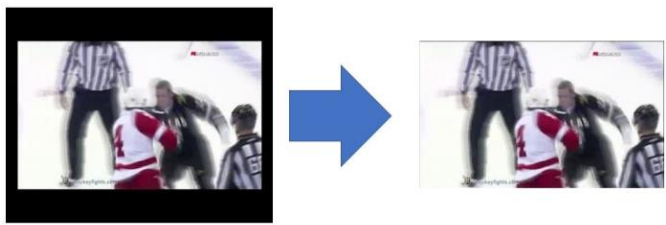

To enrich and enlarge the dataset I apply data augmentation with the transformations like image cropping and image transposition.

FIGURE 3
IMAGE CROPPING

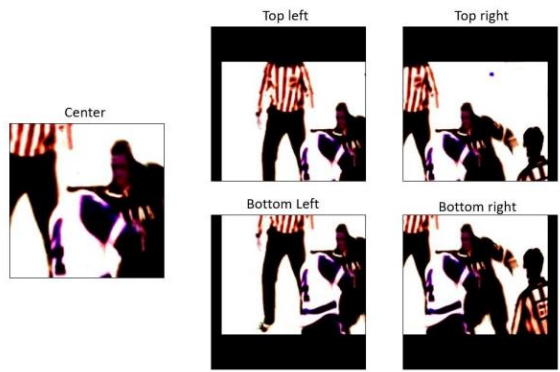

FIGURE 4
IMAGE TRANSPOSITION

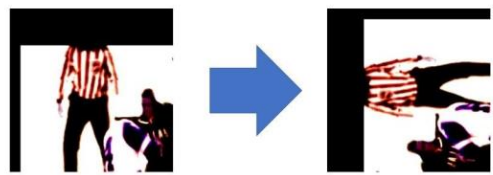

 

## ARCHITECTURE

Firstly, in the input layer, batch of 20 consecutive frames are received and their difference is computed, between every adjacent pairs. Then we use a pretrained model like RESNET50 CNN, and remove the final dense layers so it outputs the feature vector or also known as transfer values, and not the predicted class. Then these transfer values with a fixed dimension of (1,20,4096), into an LSTM layer. We also include max pooling layer of size 4 and regular dropout layers, dropping 50% information each time. At last, the chain of fully connected dense layers of size 1024, 50, 16 and finally a binary output perceptron with sigmoid activation function. We use ReLU Activation between each fully connected dense layers. To avoid internal covariate shift problem just before dense layers we apply Batch Normalization layer. I used binary cross entropy as our loss function and RMSprop as an optimizer, 20% of the data is select for validation and rest 80% is selected to train. The initial learning rate is kept 1 and is reduced by half after 5 epochs I hit plateau on validation loss.

parameters. I evaluate each hyper parameter separately and choose the best value for the next evaluations. I determined the order of the hyper-parameters to execute in a descending order of importance as follows: CNN architecture type, learning rate, sequence length, augmentation usage, dropout rate and CNN network training type (retrain or static). In Table 2, I present the different hyper parameters evaluated in each iteration.

TABLE 2
HYPER TUNING PARAMETERS

| CNN architecture | Res Net 50 | Inception V3 |
|---|---|---|
| Learning rate | 1e-4 | 1e-3 |
| Use augmentation | True | False |
| Number of frames | 20 | 30 |
| Dropout | 0 | 0.5 |
| Train type | Retrained CNN | Static CNN |

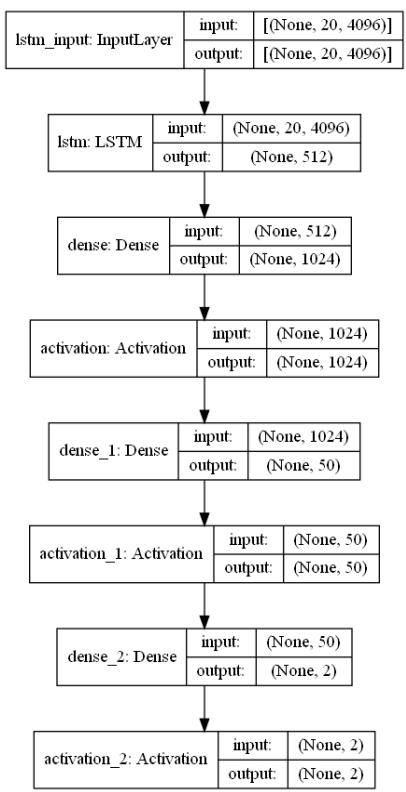

## HYPER-PARAMETER TUNING

I tried hyperparameter tuning on the Hockey dataset and then apply them for each dataset. I use only 20 epochs and early stopping of 5 instead of 15 as I apply in the final optimal network training. I also use a classic 80-20 split for train-test data. Our tuning starts with baseline hyper-

## RESULT

The hyper-tuning process as mentioned, was performed solely on the "hockey" dataset, the chosen architecture is already presented in the section. I present the hyper-tuning test accuracy for each of the hyper-parameter's values. The best performing CNN is the Resnet50[10] with 90% accuracy, the InceptionV3[11] CNN was not far from the Resnet50 with 89% accuracy but the VGG19[12] CNN had poor results of only 79% accuracy. The starting learning rate value had a critical effect on the network results where the 0.001 learning rate resolved with only 46% accuracy which is lower than the random classification. The learning rate of 0.0001 had far better results in all experiments. The augmentation increases the accuracy by 4.5% and smaller length size of the sequence improve the accuracy by 2%. the dropout of 50% did no improve the model performance and results with only 86% accuracy. As expected, where the CNN weights are not retrained had bad results of 61% accuracy.

## RESULTS PER DATASET

The results presented bellow in Figures 4, 5 and 6 are line charts with the accompanying arrangement: precision of the train (Blue), test (Gray), validation (Yellow) alongside the train loss (Orange) by number of epochs. All approaches run with 50 epochs altogether, where halting of training by the callback API of keras, occurred for all cases. The optimized model results for the" Movies" dataset presented in Figure 4 the model learning rate was reduced one time at epoch 33 and achieved 100% in training, test and validation accuracy score.

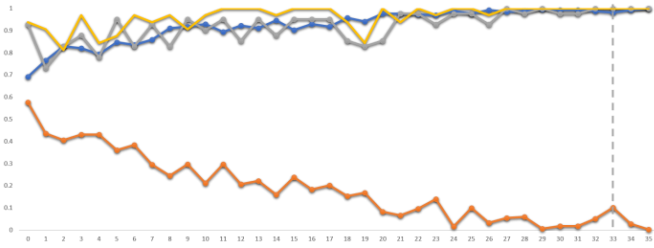

FIGURE 4
MOVIES DATASET RESULTS

As for the "Violent-Flow" dataset presented in Figure 5. the model has reduced the learning rate twice starting at 0.0001 at starting point to value of 0.00005 at the last epoch. The model test accuracy of the last epoch is 87.2% and the best overall accuracy from all of the epochs is 91.4%.

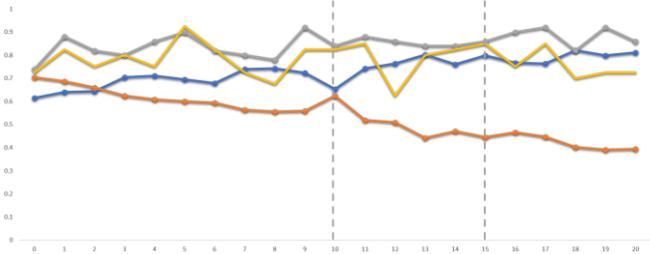

FIGURE 5
VIOLENT FLOW DATASET RESULTS

As for the "Hockey" dataset, the optimized results presented in Figure 6, the learning rate was reduced 4 times starting at 0.0001 and ending at 0.00005 at last epoch, the early stopping has stopped the training of the model at epoch 32, reaching 86.7% for the test data accuracy in the last epoch and 89.5% as the best accuracy in all of the epochs.

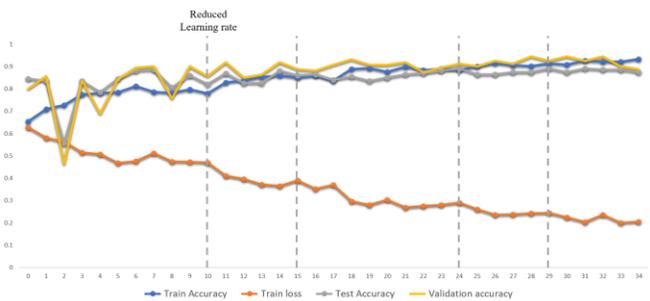

FIGURE 6
HOCKEY DATASET RESULTS

## RESULT ANALYSIS

During our evaluation I came across several hyper parameters that had critical effects on the model results. The first is the CNN architecture, the advanced Resnet50 and the inceptionV3 models had significantly better results than the VGG19. this can be easily explained by the fact that both architectures had significantly better results in image classification with much less parameters while having more network . The Resnet50 slightly outperform the inceptionV3 in the violence use case but on ImageNet the inceptionV3 is 1.5% more accurate. I believe that the depth of the Resnet50 of 168 layers compare to inceptionV3 depth of 159 might come handy in identifying the violence activity.

I found that retraining the CNN networks improves the performance of the network dramatically, the training of the CNN on the violent data help the network to tune and find relevant patterns of violence and output them to the ConvLSTM layer.

As noticed, the starting learning rate had critical effect on the learning process, the lower starting learning of 0.0001 rate prove to increase the learning of the network compare to 0.001. I assume that the high learning rate cause extreme changes of the network weights and harm its ability to converge to the right direction, the small learning rate force the network the update its weights slowly but safely into the right direction of loss.

I believe that the dropout didn't improve the network in this experiment setup because the problem and the datasets are domain specific. The need of generalization will be critical in the future cases when the video files are more heterogeneous with different video quality, camera positioning, type of scenes (not only hockey game, movies or football games) and when more classes are available such as type of violence, number of participants, violence tool, degree of injury etc.

The data augmentation process helped the model to deal with the small amount of labelled data, the augmentation increased the number of samples and helped the model to find meaningful patterns the frames. For the optimized results analysis I first dive into the" Movies" dataset, the model reached 100% accuracy. I conclude that it is a relatively "easy" dataset to classify because the learning reduction only occurred once and nearly by the end of the training session. The optimized model fitted on the" Violent-Flow" dataset had arrived at the most reduced score out of all the datasets settling at 91.4% precision. Videos in this dataset contain enormous group where even in the" violent" recordings the majority of the group is an onlooker and doesn't intercede in the brutal demonstration. one idea raised is to part the recordings into more modest pieces and to concur on a packing strategy to create the last order.

Lastly, the optimized model fitted on the" Hockey" dataset has reached 86.7%. The only dataset where the model reduced the learning rate 4 times. I suspect that in

©2021 XXXX

April 21, 2021, Mann, GJ

this dataset in particular having a long period of training the model can produce higher accuracy results, as I were limited with computing power, this could be one of the key factors explaining the 10% gap between the two models.

## FUTURE WORK

- *Drawing inference from Audio:*

    I can have an inference engine running, which uses some fuzzy logic that uses the result vector from both, video inference and audio inference, and using some composition (like min-max) draw a better inference.
    Now there are two approaches that I can think of that can help in classifying audio between two classes, violent and non-violent:
    - Raw wave input to CNN 1DConv
    - Mel Spectrogram transform input to CNN 2DConv

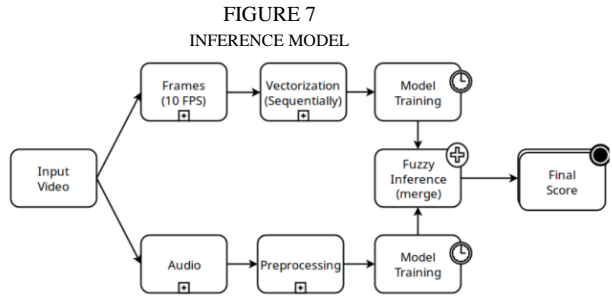

FIGURE 7
INFERENCE MODEL

Apparently, Mel Spectrogram approach has been more accurate as tested on some standard audio dataset such as Urban Sound Tag Dataset. Anyways each approach uses Short-time Fourier Transform (SFTF)[9] and it allows us to see how different frequencies change over time.

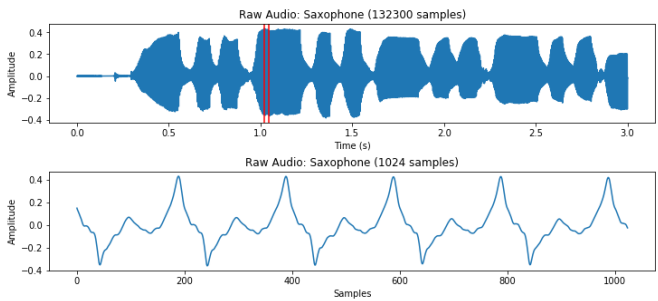

FIGURE 8
RAW WAVE INPUT

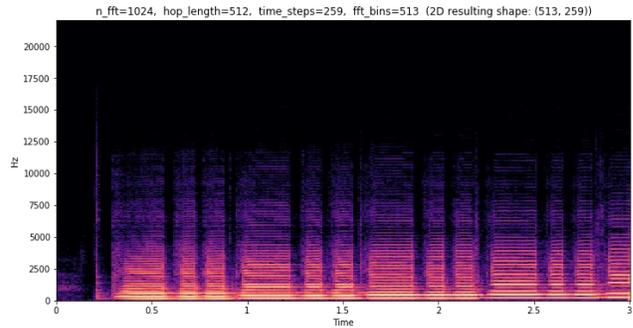

FIGURE 9
MEL SPECTROGRAM INPUT

The feature vector that can be used consist of:
• ZCR: Zero-crossing rate
• Chroma: Chromagram 1
• Chroma 2: Chromagram 2
• MFCC: Mel-frequency cepstral coefficients
• Loudness: Loudness
• Energy: Energy

And both audio and video inferences can finally be cumulated through fuzzy inference.

- *Priority based Scheduling:*

    I can devise an algorithm such that I can efficiently assign computation task to particular video stream based on the probability that in future the violence can happen.

    I devise a priority queue which has stores probability of violent activities. First, I initialize probability of each video stream to be 1/(Number of Video Streams), and then upon any detection of violent activity, I update probability by,

    $$P' = \min\left(1, P + ((isDetected) * e^{\frac{-confidence}{Number\ of\ Video\ Streams}})\right)$$

    Here, isDetected is an integer variable which can be only -1 upon no detection or 1 upon detection. The confidence coefficient is updated after drawing inference from every CCTV is just the absolute sum of difference of previous confidence and output of the LSTM.

    $$Initialize\ Confidence_i \leftarrow 0$$
    $$Confidence' = |Confidence - \sum_i output_i|$$

©2021 XXXX　　　　　　　　　　　　　　　　　　　　　　　　　　　　　　　　　　　　　April 21, 2021, Mann, GJ

And this is the scheduling queue according to which the video streams should be computed. Now the main topic of experimentation is to find the optimal time of updating the probabilities. If it's too small, the overhead of updating priority queue would add up to the delay. If the time of updating is too large, there is a possibility that a detected violent event is no taken under consideration.

I can also make a robust pipeline for this. If possible, use multi-threading to achieve partial parallelism.

*Frame Input → Preprocessing → Inference → Update Confidence*
*Frame Input → Preprocessing → Inference → Update Confidence*
*Frame Input → Preprocessing → Inference ….*

## CONCLUSION

In this work I implemented and experimented with various deepto predict violence in video data, I found our implementation to deal well with this task even though our GPU power was relatively low.

I found the smart data preprocessing of the video's frames play an important factor as well as some of the training parameters such as: CNN network, learning rate and data augmentation. Looking forward to more complex violence scenarios and appliances it will take researchers to find creative solutions for data collection, advance generalization techniques and real-time optimizations.

We also made a complementary app made in flutter for detecting violence over the CCTV or IP Webcam video streams using RTSP Protocol. Though the distributed computing cuts times in many folds still the system remains pseudo real time rather than real time while drawing inference. Though I am hopeful two-three papers down the line and we will achieve real time inference milestone, while maintaining the same precision.